%% file: paper.tex
\def\year{2019}\relax
\documentclass[letterpaper]{article} 
\usepackage{aaai19}  
\usepackage{times}  
\usepackage{helvet}  
\usepackage{courier}  
\usepackage{url}  
\usepackage{graphicx}  
\usepackage{nicefrac}       

\usepackage{mathrsfs}
\usepackage{amsmath}
\usepackage{amsfonts}
\usepackage{amsthm}
\usepackage{subfigure}
\usepackage{booktabs} 
\usepackage{wrapfig,lipsum}

\usepackage{algorithm}
\usepackage{algorithmic}

\usepackage{paralist}
\usepackage{xr}
\usepackage{color}

\input{source/defs}

\newcommand{\bs}[1]{{\boldsymbol {#1}}}

\newcommand{\minus}{\! -\! }

\newcommand{\tabref}[1]{Table~\ref{#1}}

\newcommand{\figref}[1]{Fig.~\ref{#1}}

\newcommand{\thmref}[1]{Theorem~\ref{#1}}

\newcommand{\eqnref}[1]{Eqn.~\ref{#1}}

\title{Bidirectional Inference Networks:\\A Class of Deep Bayesian Networks for Health Profiling}

\author{
Hao Wang$^1$, Chengzhi Mao$^2$, Hao He$^1$, Mingmin Zhao$^1$, Tommi S. Jaakkola$^1$, and Dina Katabi$^1$\\
\vspace{-8pt}
              \\
       $^1$MIT CSAIL, Cambridge, MA \ $^2$Columbia University, New York, NY \\
       \small \{hwang87, haohe, mingmin\}@mit.edu, cm3797@columbia.edu, \{tommi, dina\}@csail.mit.edu\\
}

\begin{document}
%

\language0
\lefthyphenmin=2
\righthyphenmin=3

\maketitle



\begin{abstract}
\input{source/abstract}
\end{abstract}

\input{source/intro}
\input{source/related}

\input{source/bin}

\input{source/cbin}

\input{source/experiment}

\input{source/conclusion}
\input{source/ack}

\bibliography{reference}
\bibliographystyle{aaai}

\end{document}

%% file: source/defs.tex
\def\A{{\bf A}}

\def\X{{\bf X}}

\def\0{{\bf 0}}
\def\1{{\bf 1}}

\def\tha{\mbox{\boldmath$\theta$\unboldmath}}

\def\argmax{\mathop{\rm argmax}}
\def\argmin{\mathop{\rm argmin}}
\newtheorem{theorem}{Theorem}

%% file: source/abstract.tex
We consider the problem of inferring the values of an arbitrary set of variables (e.g., risk of diseases) given other observed variables (e.g., symptoms and diagnosed diseases) and high-dimensional signals  (e.g., MRI images or EEG). This is a common problem in healthcare since variables of interest often differ for different patients. Existing methods including Bayesian networks and structured prediction either do not incorporate high-dimensional signals or fail to model conditional dependencies among variables. To address these issues, we propose \emph{bidirectional inference networks} (BIN), which stich together multiple probabilistic neural networks, each modeling a conditional dependency. Predictions are then made via iteratively updating variables using backpropagation (BP) to maximize corresponding posterior probability. Furthermore, we extend BIN to \emph{composite BIN} (CBIN), which involves the iterative prediction process in the training stage and improves both accuracy and computational efficiency by adaptively smoothing the optimization landscape. Experiments on synthetic and real-world datasets (a sleep study and a dermatology dataset) show that CBIN is \emph{a single model} that can achieve state-of-the-art performance and obtain better accuracy in most inference tasks than \emph{multiple models each specifically trained for a different task}.

%% file: source/intro.tex
\section{Introduction}\label{sec:intro}
In healthcare, it is often desirable to infer an arbitrary set of variables (e.g., the probability of having a particular disease) given other known or observable variables (e.g., coughing, itching, or already diagnosed diseases) and high-dimensional signals (e.g., ECG, EEG or CT)~\cite{BNlung}.  Which variables are known and which need to be inferred  typically vary across patients.  Such inference problems are important to assist clinicians in making more informed treatment decisions in clinical settings where uncertainty is ubiquitous.

Traditionally, Baysian networks (BN) are used for this task since they naturally reason with uncertain domain knowledge. However, although BN can model relationship among variables of interest, they do not model complex high-dimensional data such as MRI images or EEG time series. Deep neural networks (NN), on the other hand, are able to process such high-dimensional signals. Recently, there has been work that combines structured prediction and NN \cite{SPEN,eSPEN} to take into account the relationship among variables of interest during prediction. However, (1) they are designed to predict all variables at once without the ability to infer an arbitrary subset of variables given others; (2) they do not model conditional dependencies among variables.

In this paper, we propose bidirectional inference networks (BIN) as a kind of probabilistic NN that can get the best of all worlds: (1) it can handle high-dimensional data (e.g., EEG, CT); (2) it can infer an arbitrary subset $V_S$ of variables in $V$ given $V \setminus V_S$; (3) it can capture complex conditional dependencies among variables of interest.

Note that a naive way to infer arbitrary $V_S$ is to train different NNs for different $V_S$. However, this needs $O(2^N)$ NNs when $|V|=N$ and ignores the relationship among variables to be predicted. Instead, BIN tries to model the factorized joint distribution of $V$ by connecting multiple probabilistic NNs, each modeling a factor (conditional distribution). After it is jointly trained, predictions are made via first \emph{initializing $V_S$ with a feedforward (FF) pass} and then \emph{iteratively updating variables using backpropagation (BP) to maximize corresponding posterior probability}. BIN is then extended to CBIN, which involves the iterative prediction process in the training stage, consequently alleviating the problem of local optima and improving the performance.

We evaluate our model on two datasets (a large-scale sleep study~\cite{SHHS2} and a dermatology dataset~\cite{derm}). Experiments show that \emph{a single BIN} can predict $V_S$ given $V \setminus V_S$ with performance comparable to or better than training $O(2^N)$ separate models for $O(2^N)$ different $V_S$ when there are $N$ variables in $V$. We can also outperform state-of-the-art structured prediction models~\cite{SPEN,eSPEN} adapted for our task.

In summary, our contributions are as follows:
\begin{compactitem}
\item We propose BIN, a probabilistic neural network model that can perform bidirectional inference, handle high-dimensional data, and effectively predict an arbitrary subset $V_S$ of variables in $V$ given $V \setminus V_S$.
\item We enhance BIN with composite likelihood to derive composite BIN (CBIN), which not only improves the inference accuracy but also reduces the number of iterations needed during inference.
\item Experiments show that BIN and CBIN can outperform state-of-the-art structured prediction models adapted for our task and achieve performance comparable to or even better than training $O(2^N)$ separate models for $O(2^N)$ different $V_S$.
\item We show that it is possible to predict health status including physical and emotional well-being scores using EEG, ECG, and breathing signals.
\end{compactitem}

%% file: source/related.tex
\section{Related Work}\label{sec:related}

\subsubsection{Combination of Probabilistic Graphical Models and Deep Neural Networks}
Our work is related to a recent trend of combining probabilistic graphical models (PGM) and deep NN~\cite{BDL}. In particular, \cite{SVAE} builds on variational autoencoders (VAE)~\cite{VAE,DBLP:conf/nips/SohnLY15} to propose \emph{structured VAE} (SVAE), where they leverage conjugacy to update some variational parameters and use BP to update other parameters without conjugate structures. \cite{VSIN} generalizes SVAE to cover PGMs with non-conjugate factors and improve its performance. In addition, some methods focus on incorporating VAE into state-space models (as a kind of PGMs)~\cite{SRNN,DKF,blackbox} or recurrent neural networks (RNN)~\cite{VRNN,DRAW,STORN}. For example, deep Kalman filters \cite{DKF} use variational distributions parameterized by multi-layer perceptrons or RNN to approximate the posterior distributions of the hidden states in the state-space models. Besides the VAE-based models above, there are also models based on other forms of probabilistic NN~\cite{CDL,CKE,RDL}.

These methods focus more on generative modeling than conditional structured prediction tasks. Besides, they usually need different inference networks to infer different subsets of variables, which means that $O(2^N)$ networks are needed for general inference of N variables (each variable can be in various forms, e.g., scalars and vectors). In our work, $O(N)$ subnetworks are sufficient to support general inference of N variables. Note that one feasible way to avoid $O(2^N)$ networks in the SVAE framework is to combine it with our method to enable BP-based inference. This is used as one of our baselines. Note that our work is also different from probabilistic NNs such as \cite{NADE,MADE} (though they also model conditional distributions using NN), which fail to infer an arbitrary subset of variables and are often restricted to binary variables (see the Supplement for detailed comparison).

\subsubsection{Using Backpropagation as Inferential Procedures}
In our method, both FF and BP are involved during inference. This is different from most previous works, where only FF is involved during inference and BP is used to update parameters during training. However, the idea of using BP to (iteratively) compute predictions has been useful in some applications. For example, BP has been used to generate adversarial data points that NN would misclassify with high confidence~\cite{corr/GoodfellowSS14,corr/SzegedyZSBEGF13}, to generate embeddings for documents~\cite{PVDM}, and to perform texture synthesis or style transfer for images~\cite{NS,ControlNS}. In \cite{SPEN,eSPEN}, the authors propose variants of structured prediction energy networks (SPEN), which utilize BP to perform structured predictions. However, (1) SPEN is designed to predict all variables of interest at once given the input $\X$ and cannot perform inference on an arbitrary subset of variables given others (which is the focus of our method). (2) Unlike our method, even if SPEN is adapted to perform different inference cases, for most cases it does not have the proper `prior distributions' to provide good initialization of the target variables $V_S$ (for example, when inferring $v_2$ given $\X$, $v_1$, and $v_3$, SPEN does not have the prior $p(v_2| \X, v_1)$ to provide initialization for $v_2$ and can only rely on the general prior $p(v_2 | \X)$, which does not incorporate information of $v_1$), which is important for both accuracy and efficiency, as shown in our experiments. (3) Relationship among variables in SPEN is modeled using a global energy function while BIN models the relationship in a way similar to Bayesian networks. Hence BIN can handle conditional dependencies, more easily incorporate domain knowledge, and have better interpretability.

\subsubsection{Natural-Parameter Networks}
Our model use \emph{natural-parameter networks} (NPN) as a building block~\cite{NPN}. Different from vanilla NN which usually takes deterministic input, NPN is a probabilistic NN taking distributions as input. The input distributions go through layers of linear and nonlinear transformation to produce output distributions. In NPN, all hidden neurons and weights are also distributions expressed in closed form. As a simple example, in a vanilla linear NN $f_w(x)=wx$ takes a scalar $x$ as input and computes the output based on a scalar parameter $w$; a corresponding Gaussian NPN would assume $w$ is drawn from a Gaussian distribution $\mathcal{N}(w_m, w_s)$ and that $x$ is drawn from $\mathcal{N}(x_m, x_s)$ ($x_s$ is set to $0$ when the input is deterministic). With $\theta=(w_m,w_s)$ as a learnable parameter pair, NPN will then compute the mean and variance of the output Gaussian distribution $\mu_{\theta}(x_m, x_s)$ and $s_{\theta}(x_m, x_s)$ in closed form (bias terms are ignored for clarity) as:
\begin{align}
\mu_{\theta}(x_m,x_s)&=E[wx]=x_m w_m, \label{eq:npn_mean} \\
s_{\theta}(x_m,x_s)&=D[wx]=x_s w_s+x_s w_m^2+x_m^2 w_s, \label{eq:npn_var}
\end{align}
Hence the output of this Gaussian NPN is a tuple $(\mu_{\theta}(x_m, x_s),s_{\theta}(x_m, x_s))$ representing a Gaussian distribution instead of a single value. Input variance $x_s$ to NPN is set to $0$ in the next section. Note that since $s_{\theta}(x_m,0)=x_m^2 w_s$, $w_m$ and $w_s$ can still be learned even if $x_s=0$ for all data points (see the Supplement for generalization of NPN to handle vectors and matrices).

%% file: source/bin.tex
\section{Bidirectional Inference Networks}\label{sec:bin}
Unlike typical inference networks that perform inference by NN's feedforward pass~\cite{VAE}, our proposed bidirectional inference networks (BIN) use the feedforward pass and/or the backpropagration pass to perform inference. Such a design enables the same network to (1) predict the output given the input, and (2) infer the input from the output. In this section we introduce BIN's learning and inference process. Note that though we assume Gaussian distributions for simplicity when necessary, \emph{our framework applies to any exponential-family distributions} (see the Supplement for details).

\subsection{Notation and Motivation}\label{sec:motivation}
\subsubsection{Notation and Problem Formulation}
We use $\X$ to denote the high-dimensional information (e.g., EEG) of each subject and $V=\{v_n\}_{n=1}^{N}$ to denote the subject's $N$ attributes\footnote{Note that though in this paper $v_i$ is a scalar value, our methods are general enough to handle vectors.} of interest. For any index set $S \subseteq \{1,2\dots N\}$,
we define $V_S = \{v_n|n \in S\}$ and $V_{\minus S} = V \setminus V_S = \{v_n|n \not\in S\}$. We use $V_k$ to denote a set $\{v_n\}_{n=1}^{k}$, with $V_0 = \emptyset$. We are interested in learning a general network which is able to predict an arbitrary subset $V_S \subseteq V$ given other attributes $V_{\minus S}=V\setminus V_S$ and $\X$, the simplest case being predicting all attributes $V$ given $\X$. Input variance (corresponding to $x_s$ in Eqn.~\ref{eq:npn_mean}-\ref{eq:npn_var}) to NPN is set to $0$ in this section.

\subsubsection{Motivating Example}
Consider modeling the relations among $\X$ and variables $V_N=V_{N-1}\cup\{v_N\}$. Naively, one can learn a neural network $f(\cdot)$ with $(\X, V_{N-1})$ as input and $v_N$ as output using the loss function $\mathcal{L}=\|f(\X, V_{N-1}) - v_N\|_2^2$. Such a network has no problem predicting $v_N$ given $\X$ and $V_{N-1}$. However, if we want to infer $V_S\in V_{N-1}$ given $f(\cdot)$, $\X$, and $V_N\setminus V_S$, we need to (1) randomly initialize $V_S$, and (2) iteratively compute the gradient $\frac{\partial \mathcal{L}}{\partial V_S}$ and update $V_S$ until $\mathcal{L}$ is minimized. Unfortunately this does not work because: (1) With random initialization, $V_S$ can easily get trapped in a poor local optimum. (2) There are many adversarial gradient directions that can decrease $\mathcal{L}$ and lead $V_S$ far from the ground truth, especially in large and complex networks~\cite{mixup}.

\subsection{Model Formulation and Learning}
To alleviate the problems above, we propose to construct BIN as a `deep Bayesian network' over $V$ so that we can properly \emph{initialize} and \emph{regularize} $V_S$ for the iterative updates during inference (details of \emph{initialization} and \emph{regularization} of $V_S$ during inference will be introduced later). We first factorize the conditional joint distribution $p(V | \X)$ as:
\begin{align}
p(V | \X) = \prod_{n=1}^N{p(v_n | \X, V_{n-1})}.\label{eq:factorization}
\end{align}
Note that here we assume full factorization using the chain rule as in~\cite{NADE} only for simplicity (since usually we do not have prior knowledge on the relationship among variables, as is the case in our experiments); \emph{BIN is actually general enough to handle arbitrary factorization (i.e., Bayesian network structure)}. Besides the difference in structure flexibility, as shown in the following sections, both learning and inference of BIN are also substantially different from~\cite{NADE} and its variants. For example, BIN performs inference mainly with backpropagation, while~\cite{NADE} performs inference the usual way with feedforward; BIN naturally parameterizes each conditional with an NPN, while~\cite{NADE} uses parameter sharing to parameterize different conditionals (see the Supplement for detailed comparison). The large performance gap in Table~\ref{table:shhs8v}$\sim$\ref{table:forward_derm} also empirically verifies significance of these differences.

As mentioned above, we can naturally use $N$ NPN networks with parameters $\bs{\theta}_n$ to model each of the conditional distribution $p(v_n | \X, V_{n-1})$.
The negative joint log-likelihood for a given $V$ can be written as:
\begin{align}
 \mathcal{L}(V| \X; \bs{\theta}) = - \sum_{n=1}^N{\log{p(v_n | \X, V_{n-1}; \bs{\theta}_n)}}, \label{eqn:joint-nll}
\end{align}
where each term corresponds to an NPN subnetwork:
\begin{align}
-\log{p(v_n | \X, V_{n-1}; \bs{\theta}_n)} &= \frac{\|\mu_{\theta_n}(\X, V_{n-1}) - v_n\|_2^2}{2s_{\theta_n}(\X, V_{n-1})} \nonumber \\
&+ \frac{1}{2}\log s_{\theta_n}(\X, V_{n-1}), \label{eq:cond}
\end{align}
where we assume Gaussian NPN in \eqnref{eq:cond} and $\bs{\theta} = \{\bs{\theta}_n\}_{n=1}^N$. $\mu_{\theta_n}(\cdot)$ and $s_{\theta_n}(\cdot)$ are the output mean and variance of the $n$-th NPN (similar to Eqn.~\ref{eq:npn_mean}$\sim$\ref{eq:npn_var} with $x_s=0$). Our model is trained by minimizing the negative log-likelihood~(\eqnref{eqn:joint-nll}$\sim$\ref{eq:cond}) of all $M$ training samples.

\begin{figure*}[!tb]
\begin{center}
\subfigure{
\includegraphics[height=3.5cm]{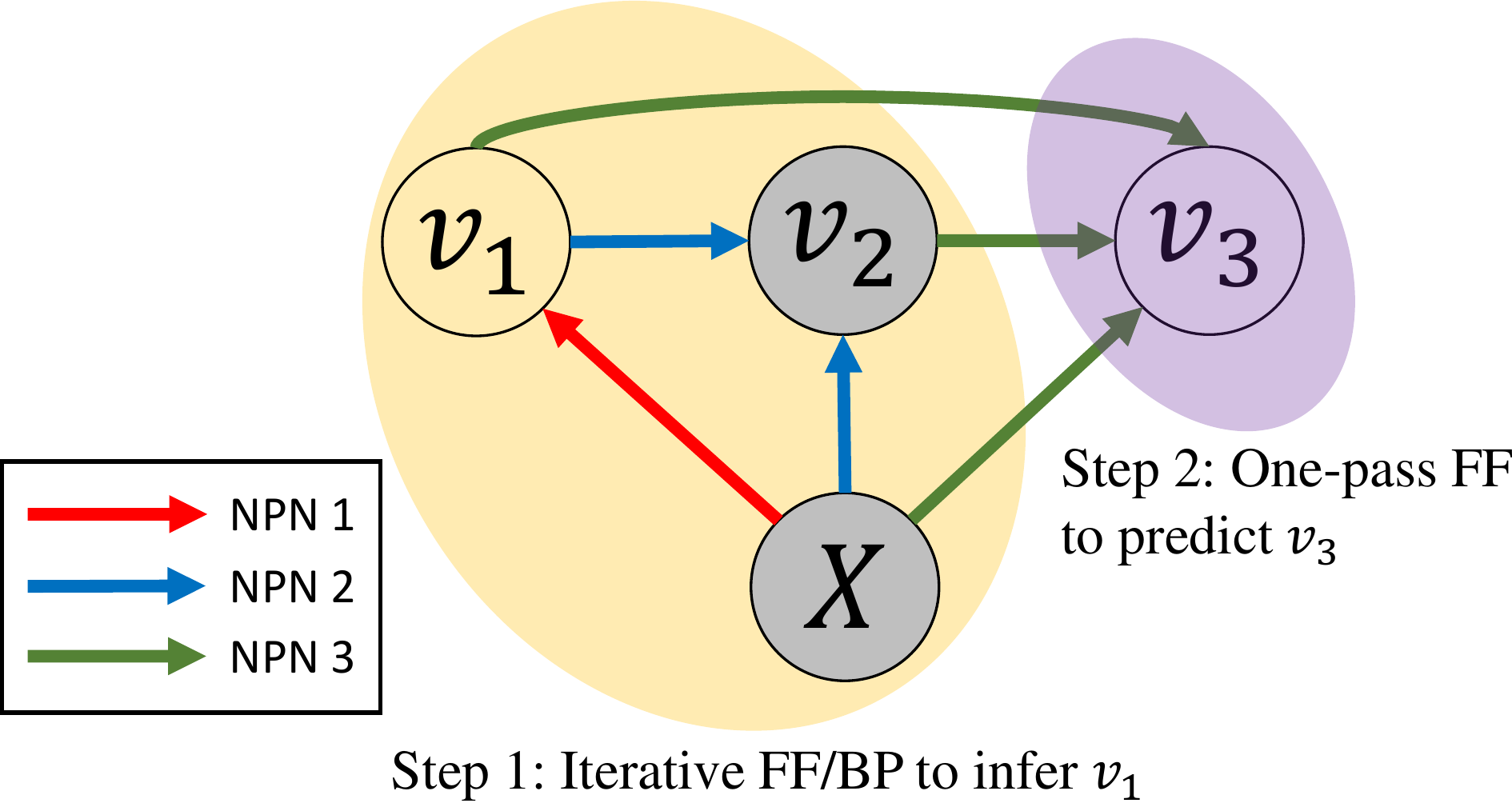}}
\hspace{-0.05in}
\subfigure{
\includegraphics[height=3.5cm]{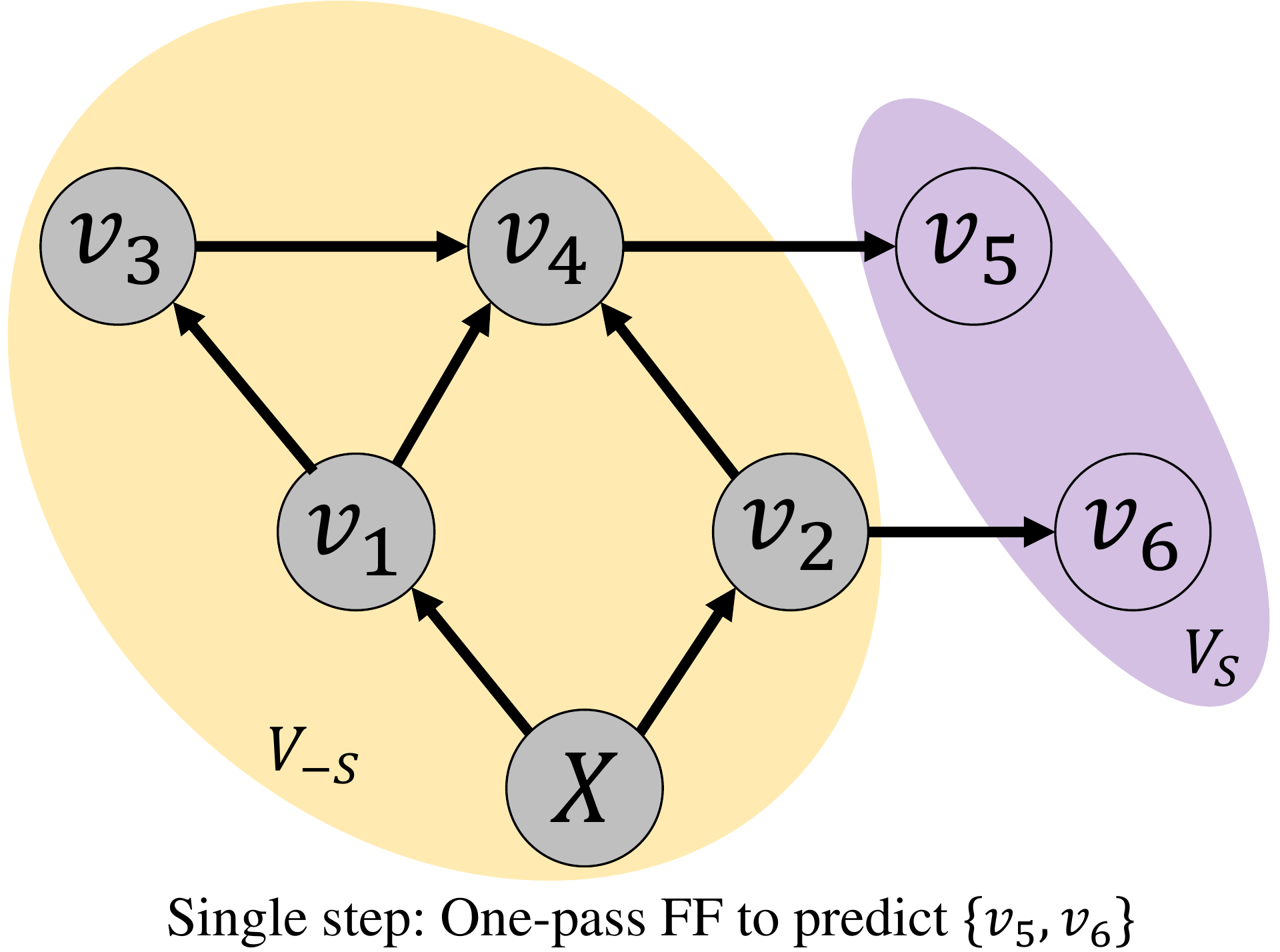}}
\hspace{-0.05in}
\subfigure{
\includegraphics[height=3.5cm]{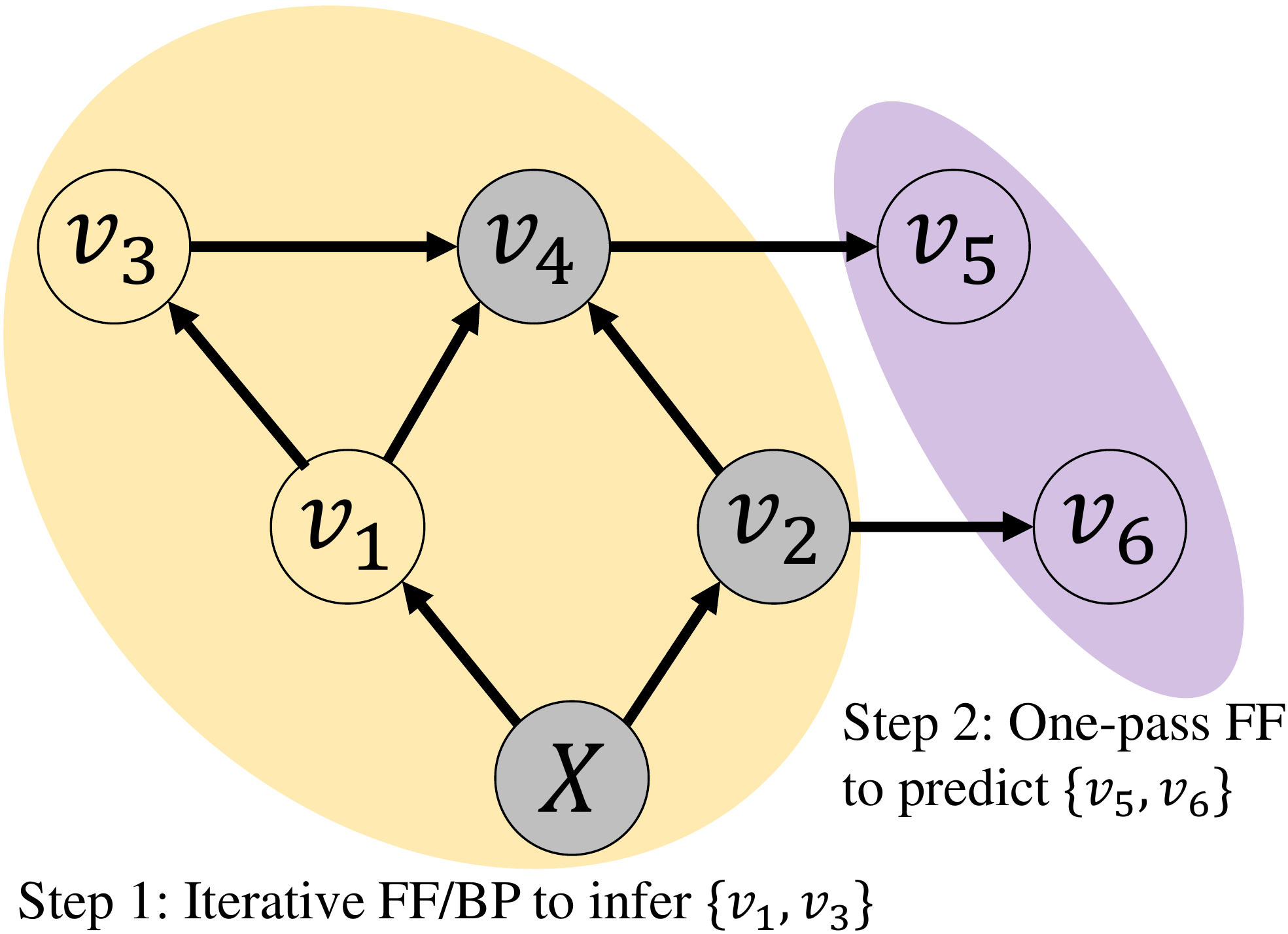}}
\end{center}
\vskip -0.20in
\caption{Transparent circles and shaded circles represent $V_S$ and $V_{\minus S}$, respectively. Left: An example for \emph{hybrid inference} when $V_S=\{v_1,v_3\}$ and $V_{\minus S}=\{v_2\}$. Edges in different colors correspond to different \emph{probabilistic} neural networks (NPN). Best viewed in color. Middle: An example for \emph{forward prediction} of a more general BN structure. Right: An example for \emph{hybrid inference} of a more general BN structure.}
\vskip -0.0in
\label{fig:bn}
\vskip -0.20in
\end{figure*}

\subsection{Inference}\label{sec:inference}
\subsubsection{General Inference}
Once the model is trained, it can be used to predict an arbitrary subset $V_S$
given all other attributes $V_{\minus S}$ and $\X$ using the maximum a posteriori (MAP) estimates of $p(V_S | \X, V_{\minus S})$:

\begin{align}\label{eq:inference}
 \argmax_{V_S}{p(V_S | \X, V_{\minus S}; \bs{\theta})}  &= \argmax_{V_S}{p(V_S, V_{\minus S}| \X; \bs{\theta})} \nonumber \\
                                                        &= \argmin_{V_S}{\mathcal{L}(V_S, V_{\minus S}| \X; \bs{\theta})},
\end{align}
where $\mathcal{L}(V_S, V_{\minus S}| \X; \bs{\theta})$ is identical to \eqnref{eqn:joint-nll}. We use Adam \cite{adam} as an adaptive gradient update procedure to minimize $\mathcal{L}(V_S, V_{\minus S}| \X; \bs{\theta})$ with respect to $V_S$ while \emph{fixing $\X$, $V_{\minus S}$ and $\bs{\theta}$}.

\textbf{On the Motivating Example}: Echoing the \emph{motivating example} in previous text, note that during the inference of $v_n \in V_S$, the term $\frac{\|\mu_{\theta_n}(\X, V_{n-1}) - v_n\|_2^2}{2s_{\theta_n}(\X, V_{n-1})}$ in \eqnref{eq:cond} provides both the \emph{initialization} (i.e., initialize $v_n$ as $\mu_{\theta_n}(\X, V_{n-1})$) and \emph{adaptive prior (regularization)} while the term $\sum_{k=n+1}^N(\frac{\|\mu_{\theta_k}(\X, V_{k-1}) - v_k\|_2^2}{2s_{\theta_k}(\X, V_{k-1})} + \frac{\log s_{\theta_k}(\X, V_{k-1})}{2})$ provides the main part of $\frac{\partial \mathcal{L}}{\partial v_n}$ to update $v_n$ (since $v_n\in V_{k-1}$ when $k>n$). The trade-off between the prior and main gradient terms then depends on predicted variance terms $s_{\theta_k}(\X, V_{k-1})$ where $k\geq n$.

\textbf{A Toy Example}: As a more concrete example, assume $N=2$ and the structure $p(v_1|\X)p(v_2|\X,v_1)$. When inferring $v_1$'s MAP estimate given $\{\X, v_2\}$: $p(v_1|\X)$ is described by \emph{sufficient statistics} $(\mu_{\theta_1}(\X),s_{\theta_1}(\X))$ produced by NPN and provides regularization $\frac{\|\mu_{\theta_1}(\X)-v_1\|_2^2}{2s_{\theta_1}(\X)}$ for $v_1$'s MAP estimate; $p(v_2|\X,v_1)$ however depends on the \emph{actual value} of $v_1$, not the NPN representation of $v_1$, and therefore also guides the MAP estimate of $v_1$ (see the Supplement for an illustrative figure on this example). Overall, MAP inference is performed by minimizing the following objective $\mathcal{L}_{toy}$ w.r.t. $v_1$ given $\{\X, v_2\}$:
\begin{align*}
\mathcal{L}_{toy} &= \frac{\|\mu_{\theta_1}(\X) - v_1\|_2^2}{2s_{\theta_1}(\X)}
+ \frac{1}{2}\log s_{\theta_1}(\X) \\
&+ \frac{\|\mu_{\theta_2}(\X, v_1) - v_2\|_2^2}{2s_{\theta_2}(\X, v_1)}
+ \frac{1}{2}\log s_{\theta_2}(\X, v_1)
\end{align*}

\subsubsection{Inference in Special Cases}
\emph{In general}, inference can be performed by iterative FF and BP to jointly find $V_S$ using \eqnref{eq:inference}. \emph{In some special cases} where $v_N\notin V_{\minus S}$, inference can be sped up by leveraging the structure of the conditional dependencies among variables. Next we consider two such cases (where $v_N\notin V_{\minus S}$):

\textbf{Case 1 (Forward Prediction)}: When the task is to predict $V_S = V \setminus V_k$ given $\X$ and $V_{\minus S}=V_k$. We can use one-pass FF to infer $V_S$ by greedy maximization as follows:
\begin{align}\label{eq:greedy}
\widehat{v}_n = \argmax_{v_n} p(v_n | \X, V_{n-1})
= \mu_{\theta_n}(\X, \widehat{V}_{n-1}),
\end{align}
where $n = k+1,\dots, N$. Note that although $\widehat{v}_n$ may not be the global optimum due to the terms $\frac{1}{2} \log s_{\theta_n}(\X, V_{n-1})$, in our preliminary experiments, we find that using $\widehat{v}_n$ as initialization and then finetuning jointly $V_S$ to minimize $\mathcal{L}$ (according to \eqnref{eq:inference}) has very similar performance. In \thmref{th:greedy_forward} below, we show that under some assumptions on $V$, our forward prediction based on greedy maximization can achieve the global optimum, which provides some insight on the similar performance.

\begin{theorem}\label{th:greedy_forward}
Assume our ground-truth joint distribution $p(V_N | \X)$ is an elliptically unimodal distribution and BIN converges to its optimal. \textbf{Forward prediction} based on greedy maximization in \eqnref{eq:greedy} achieves the global optimum.
\end{theorem}

\textbf{Case 2 (Hybrid Inference)}: When $v_N \notin V_{\minus S}$ and $V_{\minus S}\neq V_k$ for any $k$, one can first perform the (joint) general inference for variables $\{v_n | v_n \in V_S, n<q\}$, where $v_q$ is the last variable (with the largest index $q$) in $V_{\minus S}$, by \emph{iterative} FF and BP, and then perform \emph{one-pass} forward prediction for variables $\{v_n | v_n \in V_S, n>q\}$. Doing this could significantly cut down the time needed for predicting $\{v_n | v_n \in V_S, n>q\}$, since no BP and iterative process is needed for them. For example, if $V_S=\{v_1, v_3\}$ and $V_{\minus S}=\{v_2\}$, one can first perform general inference for $v_1$ and then perform forward prediction for $v_3$ (see \figref{fig:bn}(left)). Similar to \thmref{th:greedy_forward}, we show in \thmref{th:greedy_hybrid} that under some assumptions on $V$, hybrid inference can also achieve the global optimum.

\begin{theorem}\label{th:greedy_hybrid}
Assume our ground-truth joint distribution $p(V_N | \X)$ is an elliptically unimodal distribution and BIN converges to its optimal. \textbf{Hybrid inference} can achieve the global optimum for $V_S$ if the backward part achieves the global optimum for $\{v_n | v_n \in V_S, n<q\}$.
\end{theorem}

Elliptically unimodal distributions are a broad class of distributions. We provide the definition and some properties as well as the proof of the theorems above in the Supplement.

\textbf{Remark}: Empirically BIN works well even though the conditions in the theorems above do not necessarily hold. Also note that although Case 1 and Case 2 assume the BN structure in \eqnref{eq:factorization}, they can be naturally generalized to other facetorization (BN structure). For example, in a general BN structure, if there exists a topological order of variables in $V$ such that $V_{\minus S}=V_k$ (i.e., all variables in $V_S$ are descendents of $V_{\minus S}$), Case 1 can be applied similarly. See \figref{fig:bn} for some examples of both cases.

%% file: source/cbin.tex
\section{Composite Bidirectional Inference Networks}\label{sec:cbin}

\textbf{From BIN to Composite BIN}: During \emph{inference} of BIN we may need to iteratively update $V_S$ via BP. To reflect this effect during \emph{training} and tailor the model in a manner that works well with the proposed inferential procedure (to further improve the optimization landscape and speed up inference), \emph{composite BIN} (CBIN) augments \eqnref{eqn:joint-nll} with composite likelihood (CL) terms $\mathcal{L}_j$ and uses the following training objective:

\begin{align}
 \mathcal{L}_{all} &= \mathcal{L}(V| \X; \bs{\theta}) + \lambda_c \sum_{j=1}^J\mathcal{L}_j, \nonumber \\
 \mathcal{L}_j &= \mathcal{L}(\widehat{V}_{S_j}, V_{\minus {S_j}}| \X; \bs{\theta}) -  \mathcal{L}( V_{\minus {S_j}}| \X; \bs{\theta}), \label{eq:total_loss} \\
 \widehat{V}_{S_j} &\approx \argmin_{V_{S_j}}{\mathcal{L}(V_{S_j}, V_{\minus {S_j}}| \X; \bs{\theta})}, \label{eq:approx_v}
\end{align}
where $\mathcal{L}_j$ integrates the inference task of predicting $V_{S_j}$ ($S_j$ is a subset of $\{1,\dots,N\}$ specified by users) given $(\X,V_{\minus S_j})$ into the training process. $\widehat{V}_{S_j}$ is computed in an inner loop during each epoch by iteratively updating $V_{S_j}$ through BP. $\lambda_c$ is a hyperparameter. Note that \eqnref{eq:approx_v} is an approximation because the inner loop may not yield global optima. To gain more insight, $\mathcal{L}_j$ can be written as
\begin{align}
\mathcal{L}_j &= \mathcal{L}(\widehat{V}_{S_j}| \X; \bs{\theta})
+ \mathcal{L}(V_{\minus {S_j}}| \X, \widehat{V}_{S_j}; \bs{\theta}) \nonumber \\
&- \mathcal{L}(V_{\minus {S_j}}| \X; \bs{\theta}) \label{eq:decompose}, \\
\mathcal{L}(V_{\minus {S_j}}| \X; \bs{\theta}) &= -\log p(V_{\minus S_j}|\X) \nonumber \\
&= -\log \int p(V_{\minus S_j} | \X, V_{S_j}) p(V_{S_j} | \X) d V_{S_j} .\nonumber
\end{align}

\textbf{Intuition for the Objective Function}: The training process of CBIN is summarized in Algorithm \ref{alg:chbin}. Interestingly, the last two terms in \eqnref{eq:decompose} can be seen as the difference between (1) the negative log-likelihood of $V_{\minus S_j}$ given $\X$ and the inferred $\widehat{V}_{S_j}$ (which is smaller since $\widehat{V}_{S_j}$ contains information from $V_{S_j}$) and (2) the negative log-likelihood of $V_{\minus S_j}$ given only $\X$ (which is larger and serves as a baseline). Hence minimizing the last two terms makes the network aware that $\widehat{V}_{S_j}$ contains additional information and consequently decrease the loss. On the other hand, the first term in \eqnref{eq:decompose} would update the network so that the prior and initialization provided to infer $V_{S_j}$ can be closer to $\widehat{V}_{S_j}$. From another perspective, $\mathcal{L}_j$ can be seen as adapting the augmented data $(\widehat{V}_{S_j},V_{\minus S_j})$ according to the training process to make the optimization landscape of $V_{S_j}$ more friendly to inference via BP (see the next section for more details and experiments). It is also worth noting that in \eqnref{eq:total_loss}, $\mathcal{L}_j=\mathcal{L}(\widehat{V}_{S_j}|\X, V_{\minus S_j};\bs{\theta})$ can be considered as a composite likelihood (a generalized version of pseudolikelihood~\cite{PL}) term \cite{CL} given $\X$ if $\widehat{V}_{S_j}$ is replaced with $V_{S_j}$. It is proven that adding CL terms does not bias the learning of parameters \cite{CL}.

\textbf{Configuration of $V_{S_j}$}: One challenge, however, is that there are $2^N-1$ configurations of $V_{S_j}$, including all $2^N-1$ terms of $\mathcal{L}_j$ during training is obviously impractical. In our experiments, we let $J = N-1$ and $V_{S_j} = V_j = \{v_n\}_{n=1}^{j}$. Doing this has the effect of both self-correction and improving the optimization landscape: (1) \textbf{Self-correction}: The inner loop inferring $\widehat{V}_j$ given $V\setminus V_j$ can be seen as searching for the best path for correcting $\widehat{V}_j$. (2) \textbf{Optimization landscape}: The generated $\widehat{V}_j$ is used as input for $N - j$ subnetworks; hence these $N - 1$ extra terms are sufficient to cover $N$ subnetworks and improve their optimization landscape. For more details please refer to the Supplement.

\textbf{Remark}: As shown in the experiments below, the inclusion of CL terms in CBIN not only improves the accuracy and generalization but also leads to faster inference. We attribute this to the preference of CBIN to learn smoother (and consequently more generalizable) optimization landscape w.r.t. $V$ (see the optimization landscape in the Supplement and \figref{fig:regress2d} ). Note that the marginal negative log-likelihood $\mathcal{L}(V_{\minus {S_j}}| \X; \bs{\theta})$ in \eqnref{eq:total_loss} can be approximated efficiently leveraging the properties of NPN (see the Supplement for details).

In the Supplement, we prove that for single-layer NPN subnetworks, the approximation process can obtain the exact mean and variance (diagonal entries of the covariance matrix) of $p(V_{\minus S_j}|\X;\tha)$.

\begin{algorithm}
\caption{Learning CBIN}
\begin{algorithmic}[1]\label{alg:chbin}
\STATE \textbf{Input:} Data $\mathcal{D}=\{(\X^{(i)},V^{(i)}\}_{i=1}^{M}$, training iterations $T_t$, warmup iterations $T_w$, inference iterations $T_{in}$, learning rate $\rho_t$, step size $\gamma_t$, initialized model parameter $\tha=\{(\mu_{\theta_n}(\cdot),s_{\theta_n}(\cdot))\}_{n=1}^N$.
A minibatch of size $K$ for each iteration is denoted as $\{(\X^{(k)},V^{(k)})\}_{k \in \{i_1,\dots,i_K\}}$.
\FOR{$t=1:T_w$}
\STATE Update $\tha \gets \tha - \frac{\rho_t}{K} \sum_{k} \nabla_{\tha}\mathcal{L}(V^{(k)}|\X^{(k)};\tha)$ with $\mathcal{L}$ in \eqnref{eqn:joint-nll} as a loss function.
\ENDFOR
\FOR{$t=1:T_t$}
\FOR{$j=1:J$}
\STATE Initialize $\{\widehat{V}_{S_j}^{(k)}\}$ of the current minibatch for the CL term $\mathcal{L}_j$ using \eqnref{eq:greedy}.
\FOR{$t_{in}=1:T_{in}$}
\STATE Update $\{\widehat{V}_{S_j}^{(k)}\}$ via FF and BP: $\widehat{V}_{S_j}^{(k)} \gets \widehat{V}_{S_j}^{(k)} - \gamma_{t_{in}}\nabla_{\widehat{V}_{S_j}^{(k)}}\mathcal{L}(\widehat{V}_{S_j}^{(k)}, V_{-S_j}^{(k)} |\X^{(k)};\tha)$.
\ENDFOR
\ENDFOR
\STATE Update parameters: $\tha \gets \tha - \frac{\rho_t}{K} \sum_{k} \nabla_{\tha}\mathcal{L}_{all}$ with the total loss $\mathcal{L}_{all}$ defined in \eqnref{eq:total_loss}.
\ENDFOR
\end{algorithmic}
\end{algorithm}

\setlength{\textfloatsep}{0.1cm}
\setlength{\floatsep}{0.1cm}

%% file: source/experiment.tex
\section{Experiments}\label{sec:experiment}
In this section, we first compare BIN and CBIN in toy datasets to gain more insight about our models and then evaluate variants of BIN/CBIN and other state-of-the-art methods on two real-world datasets. In all tables below, we mark the best results without retraining new models in \textbf{bold} and the best results with or without retraining new models by \underline{underlining}.

\begin{figure*}[!tb]
\begin{center}
\subfigure[BIN]{
\includegraphics[height=3.4cm]{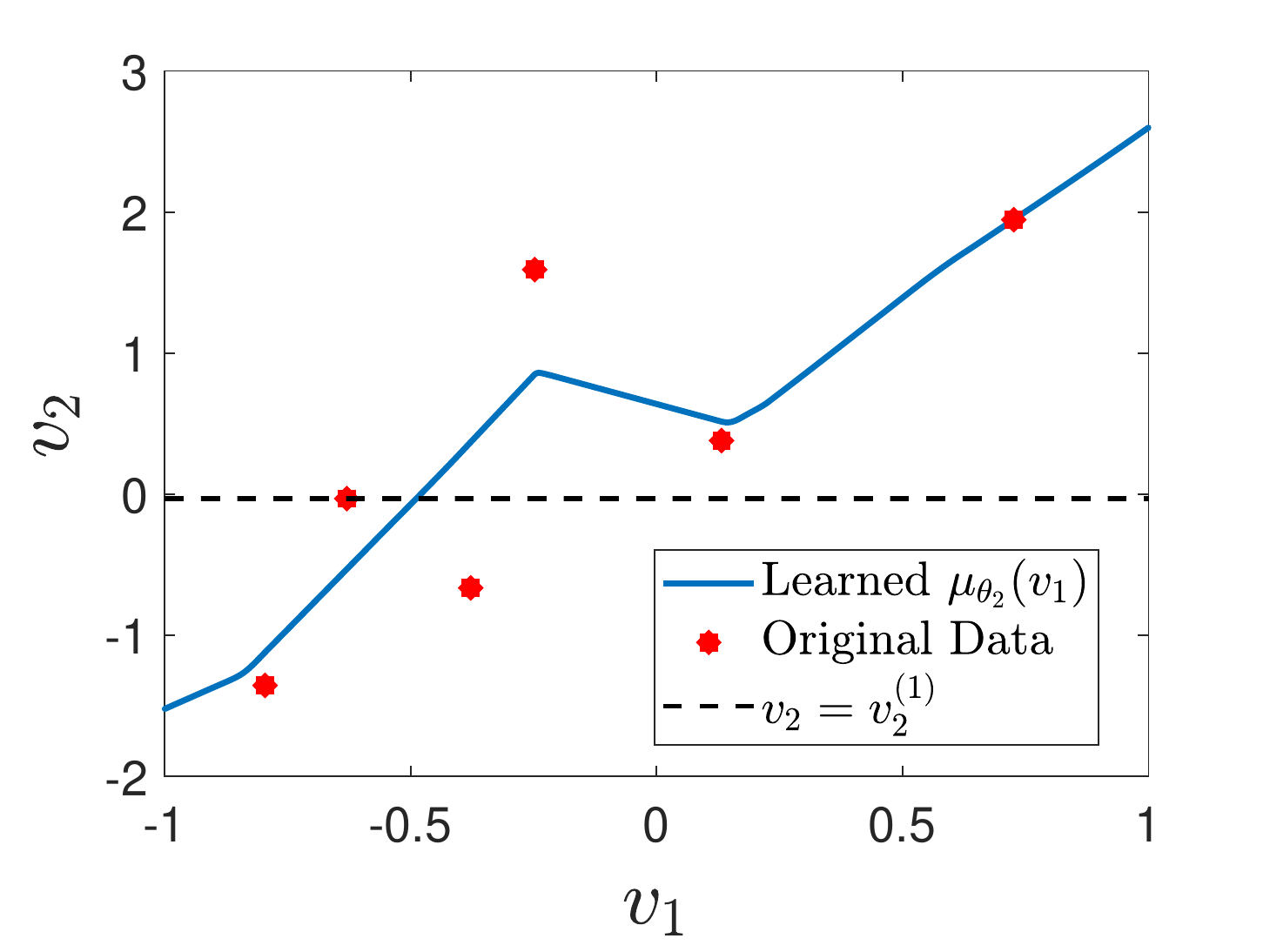}}
\hspace{-0.22in}
\subfigure[CBIN]{
\includegraphics[height=3.4cm]{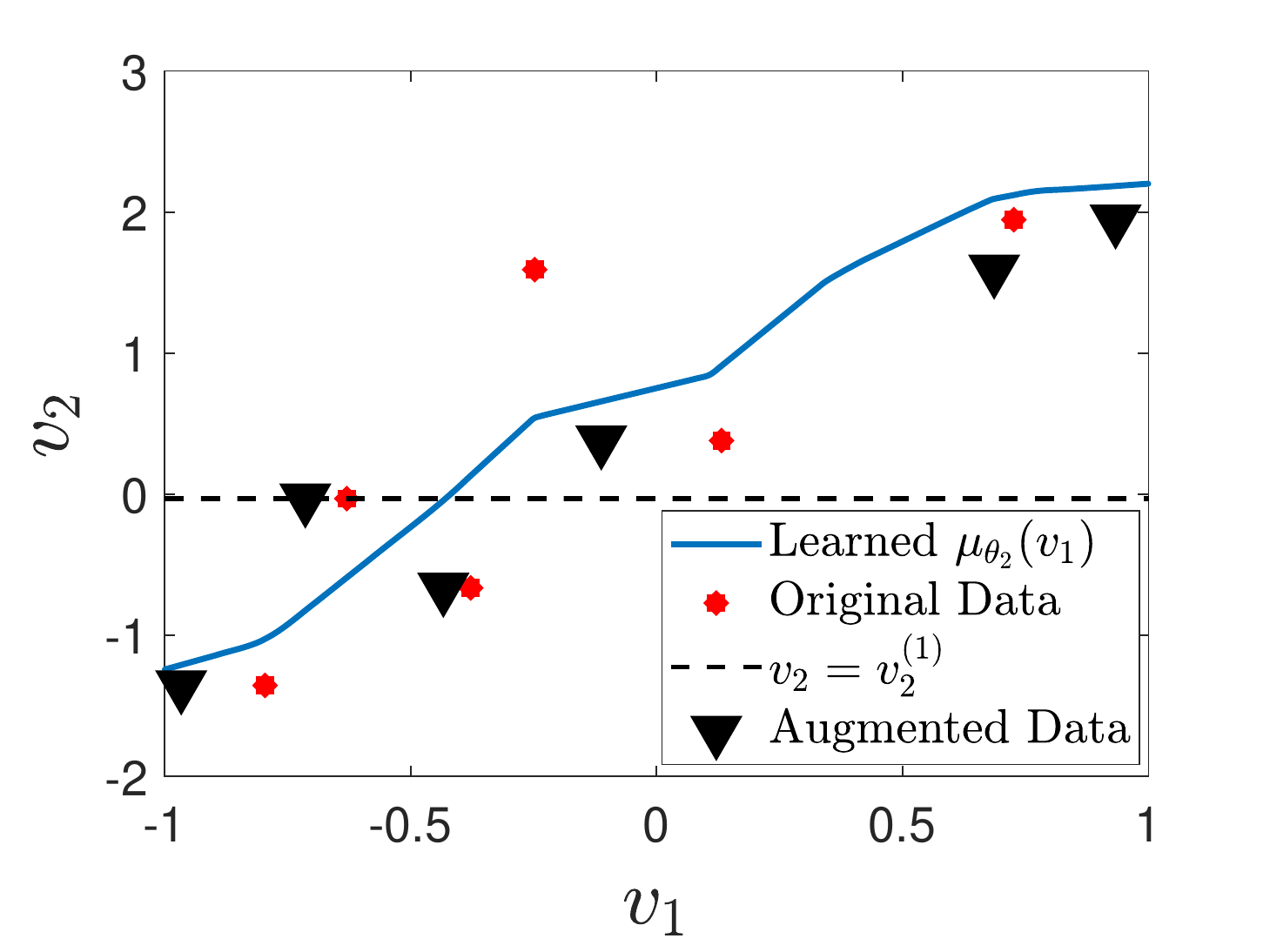}}
\hspace{-0.22in}
\subfigure[BIN]{
\includegraphics[height=3.4cm]{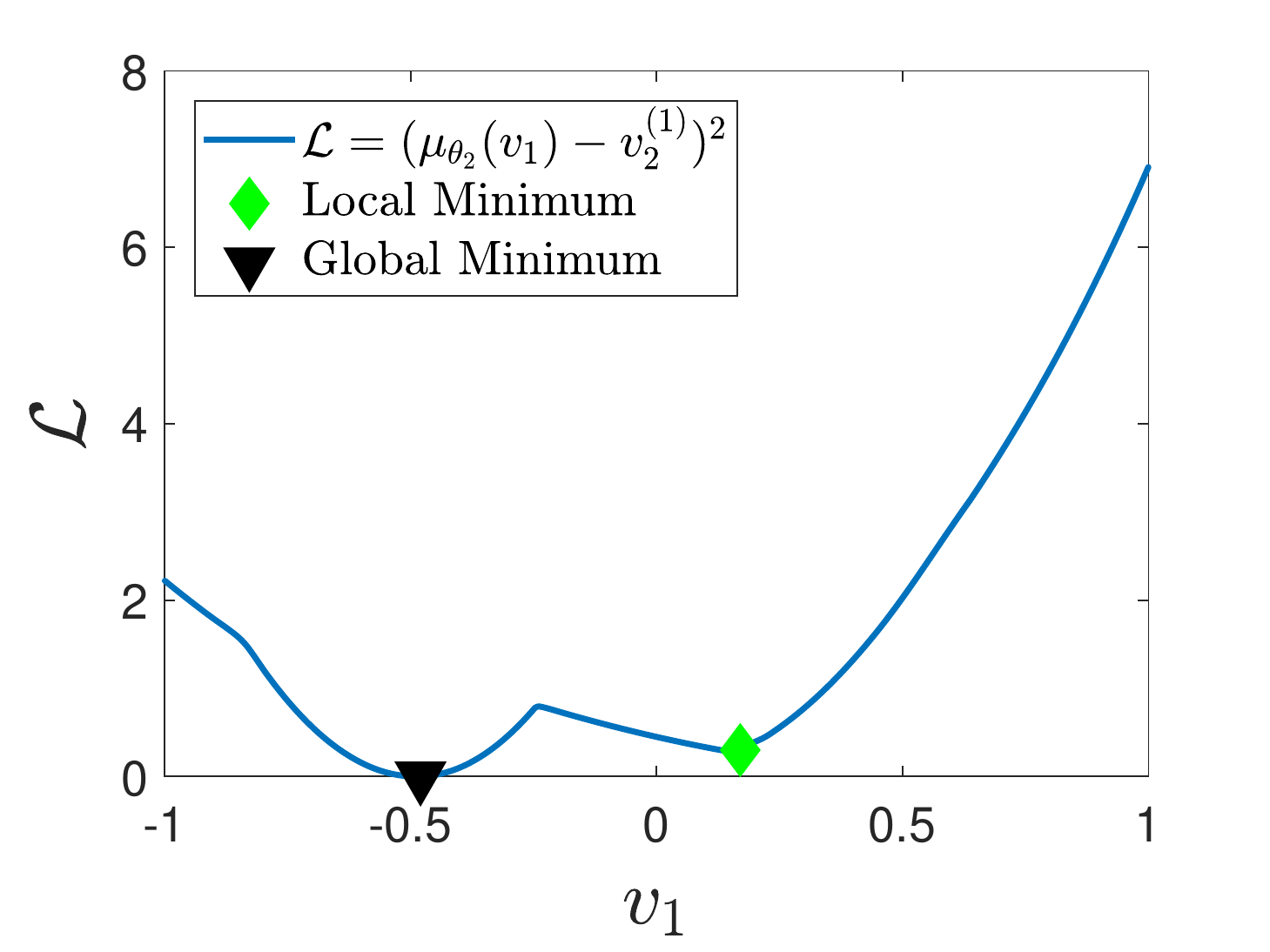}}
\hspace{-0.22in}
\subfigure[CBIN]{
\includegraphics[height=3.4cm]{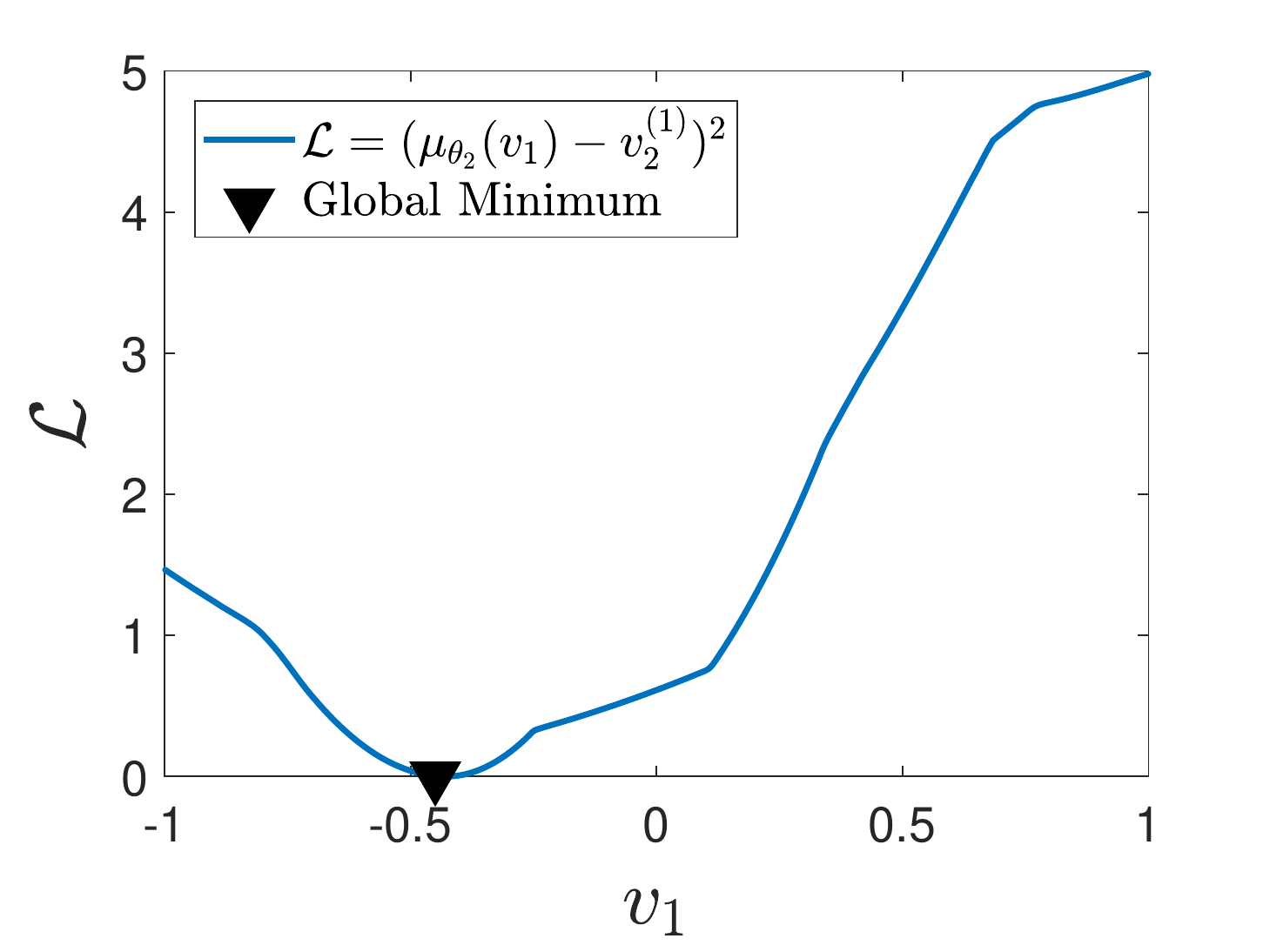}}
\end{center}
\vskip -0.25in
\caption{(a) and (b): $\mu_{\theta_2}(v_1)$ learned by BIN and CBIN. (c) and (d): corresponding loss surface of $\mathcal{L}$ with respect to $\{v_1\}$ when inferring $v_1$ given $v_2^{(1)}$ in BIN and CBIN.}
\vskip -0.0in
\label{fig:regress2d}
\vskip -0.0in
\end{figure*}

\subsection{Toy Inference Tasks}\label{sec:toy}
We start with toy datasets and toy inference tasks to show that the composite likelihood terms in CBIN help shape the function we learned to be \emph{smoother} and consequently reduce the number of local optima during inference. In the first toy dataset, $\X$ is ignored and $V=\{v_1,v_2\}$. We generate $6$ data points $\{(v_1^{(i)},v_2^{(i)})\}_{i=1}^6$ according to $v_2 = 3 v_1 + 1 + \epsilon$, where $\epsilon \sim \mathcal{N}(0, 1)$ and $v_1$ is sampled from a uniform distribution $\mathcal{U}(0, 1)$. We train BIN according to \eqnref{eqn:joint-nll} and CBIN according to \eqnref{eq:total_loss} with $J=1$ and $V_{S_1}=\{v_1\}$ (see the Supplement for details).

\figref{fig:regress2d}(a) and \figref{fig:regress2d}(b) show the $\mu_{\theta_2}(v_1)$ learned by BIN and CBIN, respectively, with the original training data points. Correspondingly, \figref{fig:regress2d}(c) and \figref{fig:regress2d}(d) show the loss surface of $\mathcal{L}$ with respect to $V_S=\{v_1\}$ when inferring $v_1$ given $v_2^{(1)}$ ($v_2^{(1)}$ corresponds to the dashed line). During inference, BIN searches for the lowest point in the loss surface of \figref{fig:regress2d}(c) and may be trapped in the poor local optimum (e.g., the diamond point in \figref{fig:regress2d}(c)). In contrast, CBIN can alleviate this problem, as shown in \figref{fig:regress2d}(d), where there is no poor local optimum. It would be interesting to inspect $\widehat{v}_1$ generated in the inner loop when training CBIN.  \figref{fig:regress2d}(b) plots augmented points $(\widehat{v}_1,v_2)$ generated in the last inner loop iteration as triangles. As can be seen, the augmented points concentrate near a straight line and hence help learn a smoother function $\mu_{\theta_2}(v_1)$ from $v_1$ to $v_2$, leading to a \emph{smoother loss surface} that is more friendly to gradient-based inference (shown in \figref{fig:regress2d}(d)). Besides this toy inference task, we also examine a more complex toy dataset where $\X$ is also considered and obtain similar results (see the Supplement for details).

\subsection{Experiments on the \emph{SHHS2} Dataset}\label{sec:shhs}

\begin{table}[!tb]
\centering
\vskip -0.5cm
\caption{$8$ variables in the \emph{SHHS2} dataset.}\label{table:sf}
\vskip 0.09cm
\begin{tabular}{c|l} \hline
$v_1$   & Physical functioning  \\ \hline
$v_2$   & Role limitation due to physical health  \\ \hline
$v_3$   & General health  \\ \hline
$v_4$   & Role limitation due to emotional problems  \\ \hline
$v_5$   & Energy/fatigue  \\ \hline
$v_6$   & Emotional well being  \\ \hline
$v_7$   & Social functioning  \\ \hline
$v_8$   & Pain  \\ \hline
\end{tabular}
\vskip 0.2cm
\end{table}

\begin{table*}[!tb]
\begin{small}
\centering
\vskip -0.3cm
\caption{Accuracy (\%) for predicting $V_S$ given $\X$ and $V_{\minus S}=\{v_n\}_{n=1}^8 \setminus V_S$ in the \emph{SHHS2} dataset. }\label{table:shhs8v}
\vskip 0.05cm
\begin{tabular}{l|ccccccc} \hline
$V_S$   & $\{v_1,v_3\}$ & $\{v_4,v_5\}$ & $\{v_1, v_3, v_6, v_7\}$ & $\{v_2, v_6, v_7\}$ & $\{v_3, v_5, v_8\}$ & $\{v_4, v_5, v_6\}$ & $\{v_4, v_6, v_7\}$\\ \hline
SPEN & 63.78 & 70.21 & 64.29 & 65.72 & 59.84 & 66.51 & 70.26 \\ 
eSPEN & 64.39 & 71.13 & 64.22 & 65.63 & 61.17 & 66.92 & 69.95 \\ 
SVAE & 62.07 & 69.59 & 62.51 & 65.13 & 59.73 & 66.03 & 68.32  \\ 
DNADE & 69.83 & 74.32 & 67.58 & 68.88 & 64.77 & 68.15 & 72.39  \\ 
PO &	68.39 & 76.29 & 70.98 & 75.00 & 71.52 & 69.75 & 74.00 \\ 
RI  &	68.08 & 68.06 & 66.26 & 67.22 & 63.48 & 65.80 & 66.62 \\ 
BIN 	& 75.31 & 79.07 & 73.48 & 75.00 & 72.55 & \underline{\textbf{74.36}} & 76.11 \\ 
CBIN  &	\underline{\textbf{77.16}}	& \underline{\textbf{80.22}} & \underline{\textbf{75.21}} & \textbf{75.87} & \underline{\textbf{72.68}} & 73.85 & \underline{\textbf{76.31}} \\ \hline \hline
Retrain  & 75.92 & 79.65 & 74.94 & \underline{76.04} & 72.24 & 73.58 & 75.74  \\ \hline
\end{tabular}
\vskip -0.55cm
\end{small}
\end{table*}

Besides synthetic datasets, we also evaluate BIN and CBIN on the sleep heart health study 2 (\emph{SHHS2}) dataset. \emph{SHHS2} contains full-night Polysomnography (PSG) from $2{,}651$ subjects. Available PSG signals include Electroencephalography (EEG), Electrocardiography (ECG), and breathing signals (airflow, abdomen, and thorax). For each subject, the dataset also includes the 36-Item Short Form Health Survey (SF-36)~\cite{SF36}. SF-36 is a standard survey that is widely used to extract 8 health variables (as shown in \tabref{table:sf}). Each of the 8 variables is represented using a score in $[0, 100]$. In the experiments, we consider PSG as high-dimensional information $\X$ and the $8$ scores $\{v_n\}_{n=1}^8$ as attributes of interest. Since the scores are based on subjects' self-reported results and intrinsically noisy, preprocessing is necessary. In particular, we use the mean of each score over all subjects as the threshold to binarize the scores into $\{0, 1\}$, where $0$ indicates `unhealthy' and $1$ indicates `healthy'. Here $\X$ is dense, high-dimensional, variable-length signals that consist of EEG spectrograms $\X_e\in \mathbb{R}^{64\times l^{(i)}}$, ECG spectrograms $\X_c\in \mathbb{R}^{22\times l^{(i)}}$, and breathing $\X_b\in \mathbb{R}^{3\times 10l^{(i)}}$, where $l^{(i)}$ is the number of seconds for the $i$-th subject (in the range $7278\sim 45448$).

We compare our models with the following baselines: \textbf{`Prior Only' (PO)} refers to using only the prior part of the trained BIN to directly output predictions without iterative inference during testing (e.g., use $\mu_{\theta_1}(\X)$ from the first subnetwork as predictions when $V_S= \{v_1\}$ and $V_{\minus S}=\{v_2\}$). \textbf{`Random Initialization' (RI)} refers to randomly initializing $V_S$ before iterative inference instead of using the prior's output during testing (e.g., during the inference for $V_S = \{v_1\}$ given $V_{\minus S}\{v_2\}$, use random initialization rather than $\mu_{\theta_1}(\X)$). \textbf{SPEN} refers to the structured prediction energy networks and \textbf{eSPEN} is its end-to-end variant~\cite{SPEN,eSPEN}. \textbf{SVAE} refers to combining SVAE~\cite{SVAE,VAE} and our method to enable BP-based inference and avoid $O(2^N)$ networks (see the Supplement for details). \textbf{DNADE} is the orderless and deep Neural Autoregressive Distribution Estimation (NADE)~\cite{NADE} proposed in~\cite{DNADE}. It is combined with the real-valued NADE~\cite{RNADE,DNADE} for the regression task. \textbf{`Retrain'} means retraining a model for that specific inference task (i.e., retraining an end-to-end NPN with $\X$ and $V_{\minus S}$ as input and $V_S$ as output). Note that the original SPEN and eSPEN \emph{can only predict $V$ given $\X$}. We adapt them for different inference tasks by resetting $V_{\minus S}$ to the given values in each inference iteration (essentially learn a density estimator $f(V,\X)$ and optimize $\max_{V_S} f(V_S,V_{\minus S},\X)$). Please refer to the Supplement for details on hyperparameters and model training of BIN, CBIN, and all baselines.

\begin{table}[!tb]
\vskip -0.4cm
      \centering
        \caption{RMSE for predicting $V_S$ given $\X$ and $V_{\minus S}=\{v_n\}_{n=1}^3\setminus V_S$ in the \emph{Dermatology} dataset. }\label{table:derm3v}
            \begin{tabular}{l|cccc} \hline
            $V_S$   & $\{v_1\}$ & $\{v_2\}$ & $\{v_1, v_2\}$ & $\{v_1, v_3\}$  \\ \hline
            SPEN &	0.0973 & 0.1310 & 0.1163 & 0.2464 \\ 
            eSPEN &	0.0944 & 0.1243 & 0.1138 & 0.2401 \\ 
            SVAE &	0.0998 & 0.1373 & 0.1236 & 0.2489 \\ 
            DNADE &	0.0828 & 0.1179 & 0.1122 & 0.2321 \\ 
            PO &	0.0979 & 0.1071 & 0.1113 & 0.2395 \\ 
            RI  & 0.0787 & 0.1301 & 0.1248 & 0.2333 \\ 
            BIN 	&	0.0691 & 0.1087 & 0.1069 & 0.2292 \\ 
            CBIN  &	\underline{\textbf{0.0643}}	& \textbf{0.1062} & \underline{\textbf{0.1011}} & \underline{\textbf{0.2130}} \\ \hline \hline
            Retrain  &	0.0714	& \underline{0.1059} & 0.1058 & 0.2271  \\ \hline
            \end{tabular}
    \hspace{1.0em}
    \vskip 0.1cm
\end{table}

\begin{table}[!tb]
\vskip -0.3cm
    \hspace{1.0em}
      \centering
        \caption{RMSE when $V_S=V$ for the \emph{Dermatology} dataset. }\label{table:forward_derm}
            \begin{tabular}{l|ccc} \hline
            $V_S$  & $\{v_1\}$ & $\{v_1, v_2\}$ & $\{v_1, v_2, v_3\}$  \\ \hline
            SPEN & -	& 0.1135 & 0.2148  \\ 
            eSPEN & -	& 0.1120 & 0.2109  \\ 
            SVAE & -	& 0.1132 & 0.2086  \\ 
            DNADE & -	& 0.1185 & 0.2161  \\ 
            BIN & -	& 0.1118 & 0.2010  \\ 
            CBIN & - &	\underline{\textbf{0.1098}}	& \underline{\textbf{0.1967}}   \\ \hline \hline
            Retrain & \underline{0.0950} &	0.1144	& 0.2059  \\ \hline
            \end{tabular}
    \vskip 0.2cm
\end{table}

\tabref{table:shhs8v} shows the accuracy of predicting different $V_S$ given $V_{\minus S}$ when $V=V_8$ (see the Supplement for additional results when $V=V_3$) for BIN, CBIN, and the baselines, where the accuracy is averaged across all inferred variable. As we can see: (1) BIN significantly outperforms `Prior Only', verifying the effectiveness of the iterative inference process. (2) BIN also outperforms `Random Initialization', which verifies the effectiveness of the initialization provided by the prior part (e.g., $\mu_{\theta_1}(\X)$ when $V_S = \{v_1\}$ and $V_{\minus S}=\{v_2\}$). (3) Furthermore, CBIN consistently outperforms BIN, which means the CL terms (and inner loops during training) are helpful to shape the optimization landscape with respect to $V_S$ so that the iterative inference process reaches better local optima (or even global optima). (4) Interestingly in most cases, CBIN can even outperform a new model trained for the specific $V_S$ and $V_{\minus S}$ possibly because CBIN (and BIN) can take into account the conditional dependency among different variables while the specifically retrained model cannot. (5) SVAE, DNADE, SPEN, and eSPEN perform poorly since they are not designed for arbitrary inference tasks or fail to properly model conditional dependencies. Besides, we also compare the accuracy of different methods in \emph{forward prediction} cases where $V_S = V$ with different $V$. We find that SVAE, SPEN, and eSPEN achieve similar or slightly better accuracy than the retrained specific models while BIN and CBIN outperform all the above baselines (see the Supplement for more experimental results).

\figref{fig:sensitivity}(left) shows the number of inference iterations needed to predict $V_S=\{v_1,v_2\}$ given $V_{\minus S}=\{v_3\}$ versus number of inner loop iterations during training ($T_{in}$ in Algorithm \ref{alg:chbin} of the main paper) with different $\lambda_c$. \figref{fig:sensitivity}(right) shows the corresponding accuracy versus $T_{in}$. The horizontal lines show the number of inference iterations and accuracy for BIN. As we can see: (1) CBIN needs much fewer iterations during testing if $T_{in}$ is large enough to get better estimates of $V_S$ during training. (2) CBIN consistently outperforms BIN in a wide range of $T_{in}$. Results for other $V_S$ (and $V$) are consistent with \figref{fig:sensitivity}.

\begin{figure}[!tb]
\begin{center}
\vskip -0.0in
\subfigure{
\includegraphics[height=3.2cm]{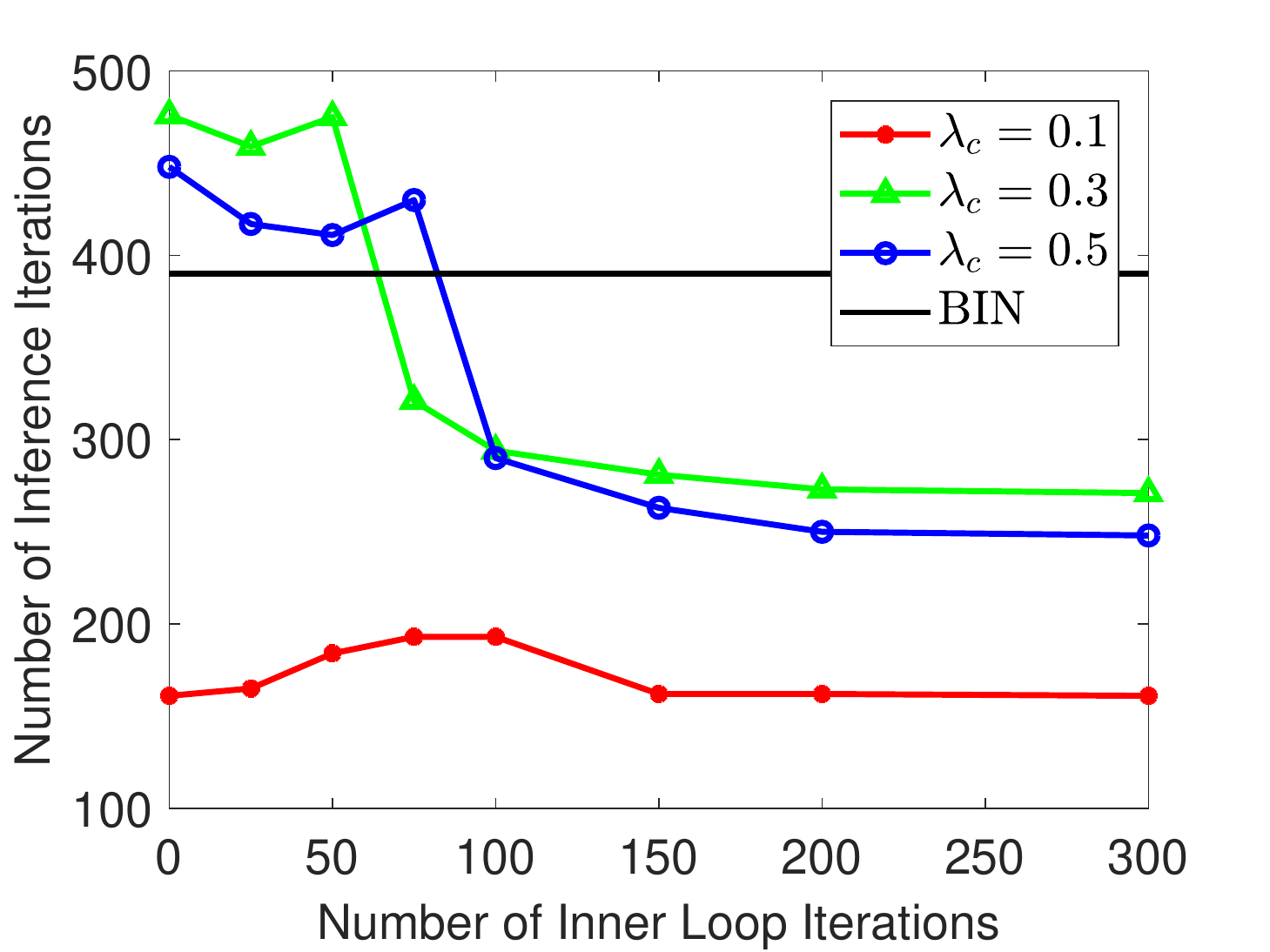}}
\hspace{-0.23in}
\subfigure{
\includegraphics[height=3.2cm]{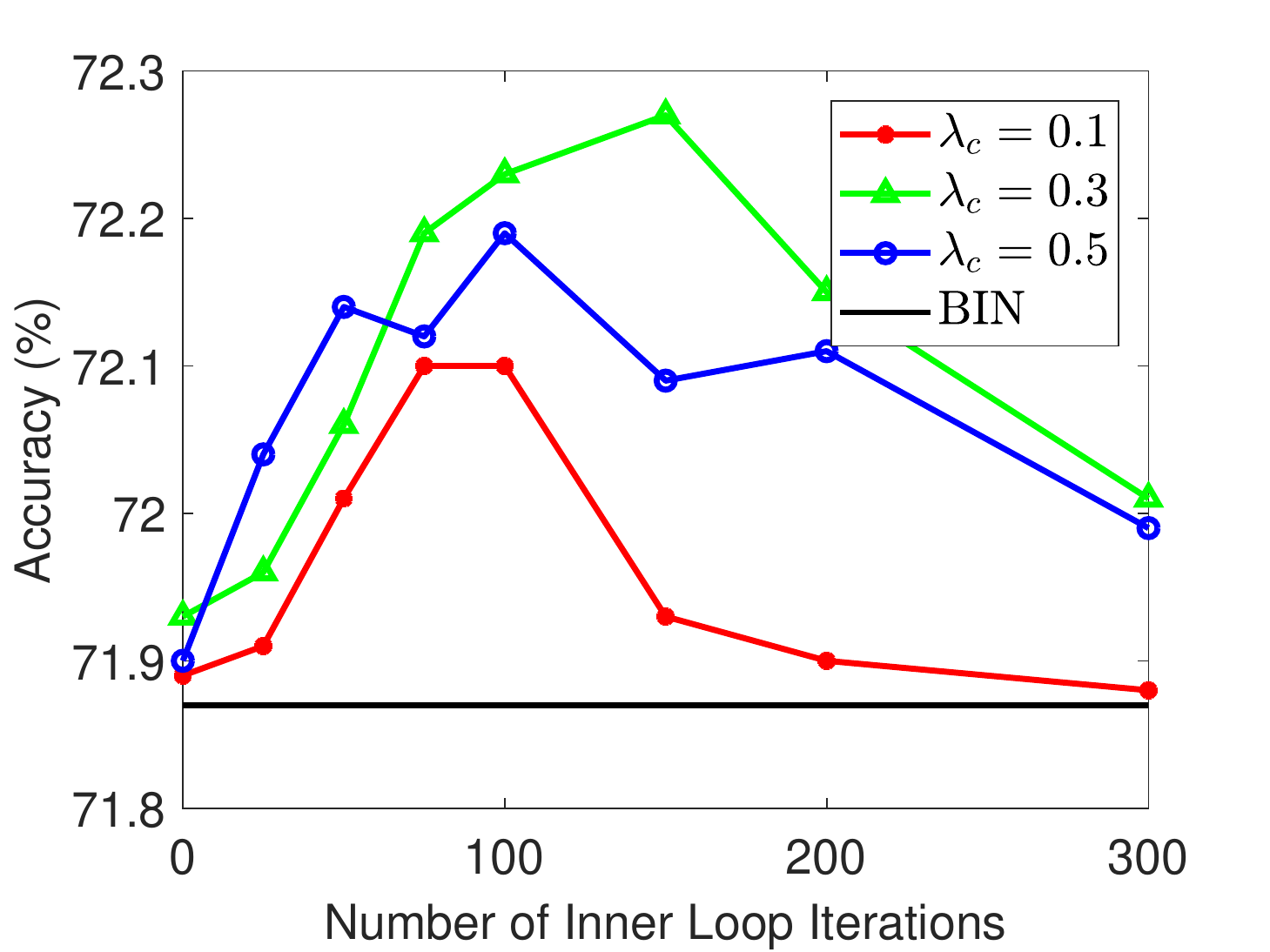}}
\end{center}
\vskip -0.25in
\caption{Left: Number of inference iterations needed during testing versus number of inner loop iterations during training ($T_{in}$ in Algorithm \ref{alg:chbin}) with different $\lambda_c$. The horizontal line the number for BIN (without $\mathcal{L}_j$). Right: Accuracy versus $T_{in}$. Similarly the horizontal line shows the accuracy of the corresponding BIN.
}
\vskip -0.0in
\label{fig:sensitivity}
\vskip 0.1in
\end{figure}

\subsection{Experiments on the \emph{Dermatology} Dataset}
Besides classification tasks, we also evaluate our methods on regression tasks using the \emph{Dermatology} dataset, which contains $12$ clinical features (e.g., itching, erythema, age, etc.) and $21$ histopathological attributes (e.g., saw-tooth appearance of retes) \cite{derm} of $366$ subjects. In the experiments, we use all $12$ clinical features as $\X$ with `vacuolisation and damage of basal layer', `saw-tooth appearance of retes', and `elongation of the rete ridges' as attributes of interest ($v_1$, $v_2$, and $v_3$). For details on hyperparameters and model training, please refer to the Supplement.

Similar to the \emph{SHHS2} experiments, \tabref{table:derm3v} shows the Root Mean Square Error (RMSE) of predicting $v_1$, $v_2$, and $v_3$ with different $V_S$ for BIN, CBIN, baselines, and retraining a model for that specific inference task. \tabref{table:forward_derm} shows the RMSE in \emph{forward prediction} cases where $V_S = V$ with different $V$. The results and conclusions are consistent with those in the \emph{SHHS2} experiments.

%% file: source/conclusion.tex
\section{Conclusion}
In this paper, we propose BIN to connect multiple probabilistic neural networks in an organized way so that each network models a conditional dependency among variables. We further extend BIN to CBIN, involving the iterative inference process in the training stage and improving both accuracy and computational efficiency. Experiments on real-world healthcare datasets demonstrate that BIN/CBIN can achieve state-of-the-art performance in the arbitrary inference tasks with \emph{a single model}. As future work it would be interesting to evaluate different factorizations for the joint likelihood of variables and different distributions (e.g., gamma distributions) beyond Gaussians. It would also be interesting to extend the models to handle hidden variables and missing values.

%% file: source/ack.tex
\section{Acknowledgments}
The authors thank Xingjian Shi, Vikas Garg, Jonas Mueller, Hongyi Zhang, Guang-He Lee, Yonglong Tian, other members of NETMIT and CSAIL, and the reviewers for their insightful comments and helpful discussion.